\pdfoutput=1
\documentclass[11pt]{article}

\usepackage[]{acl}

\usepackage{times}
\usepackage{latexsym}

\usepackage[T1]{fontenc}
\usepackage[utf8]{inputenc}

\usepackage{microtype}

\usepackage{qtree}[center]
\usepackage{forest}
\usepackage{enumitem}
\usepackage{amsmath}
\usepackage{amssymb}
\usepackage{stmaryrd}
\usepackage{linguex}
\usepackage{relsize}
\usepackage{tabularx}
\usepackage{booktabs}
\usepackage{multirow}
\usetikzlibrary{arrows.meta,matrix,positioning}
\usepackage{caption}
\usepackage{subcaption}

\newcommand{\nts}{\ensuremath{\mathcal{N}}}
\newcommand{\nt}[1]{\ensuremath{\textsc{\lowercase{#1}}}}
\newcommand{\ns}{\ensuremath{\nts_{n}}}
\newcommand{\vs}{\ensuremath{\nts_{v}}}
\newcommand{\rules}{\mathcal{R}}

\newcommand{\mcfgrule}[2]{\ensuremath{#1 & \quad\longleftarrow\quad #2 &}}
\newcolumntype{R}{>{\raggedleft\arraybackslash}X}
\newcolumntype{L}{>{\raggedright\arraybackslash}X}
\newcolumntype{C}{>{\centering\arraybackslash}X}
\newcommand\nr[2]{\ensuremath{\scriptstyle #1}}

\title{Discontinuous Constituency and BERT:\\A Case Study of Dutch}

\author{Konstantinos Kogkalidis\thanks{\quad Equal contribution.} \and Gijs Wijnholds\footnotemark[1] \\
		Utrecht Institute of Linguistics OTS, Utrecht University \\
  	    \texttt{k.kogkalidis,g.j.wijnholds@uu.nl}}

\begin{document}
\maketitle

% CONFIG
\abovedisplayshortskip=0pt              % smaller display math mode spacing
\belowdisplayshortskip=0pt              % ditto
\abovedisplayskip=0pt                   % ditto
\belowdisplayskip=0pt                   % ditto
\setlength{\Exlabelsep}{0em}            % smaller tier 1 margins for linguex
\setlength{\SubExleftmargin}{1.1em}     % smaller tier 2 margins for linguex
\setlength{\Extopsep}{.33\baselineskip} % smaller vertical skip for linguex
\renewcommand{\firstrefdash}{}          % remove dash from subexample references
\let\eachwordone=\rm                    % 1st line gloss style            
\let\eachwordtwo=\sf                    % 2nd line gloss style
\renewcommand{\trans}[1]{#1}

\begin{abstract}
In this paper, we set out to quantify the syntactic capacity of BERT in the evaluation regime of non-context free patterns, as occurring in Dutch.
We devise a test suite based on a mildly context-sensitive formalism, from which we derive grammars that capture the linguistic phenomena of control verb nesting and verb raising.
The grammars, paired with a small lexicon, provide us with a large collection of naturalistic utterances, annotated with verb-subject pairings, that serve as the evaluation test bed for an attention-based span selection probe.
Our results, backed by extensive analysis, suggest that the models investigated fail in the implicit acquisition of the dependencies examined.
\end{abstract}

\section{Introduction}
Assessing the ability of large-scale language models to automatically acquire aspects of linguistic theory has become a prominent theme in the literature ever since the inception of BERT~\cite{devlin-etal-2019-bert} and its many variants, largely due to their unanticipated performance.
Standard practice involves attaching BERT to a shallow neural model of low parametric complexity, and training the latter at detecting various linguistic patterns of interest, revealing in the process the amount to which they are encoded within BERT's representations.
The consensus points to BERT-like models having some capacity for syntactic understanding~\cite{rogers-etal-2020-primer}.
Their contextualized representations encode structural hierarchies~\cite{lin2019open} that can be projected into parse structures, using linear~\cite{hewitt-manning-2019-structural} or hyperbolic transformations~\cite{chen2021probing}, from which one can even obtain an accurate reconstruction of the underlying constituency tree~\cite{vilares2020parsing}.

Despite their broadening scope, a latent bias persists in the insights provided by the probing literature, due to its focus being, by default, on English.
English, albeit boasting a rich collection of evaluation resources, is characterized by a simple grammar with relatively few complications over the syntactic and morphological axes.
Specifically when it comes to syntax, English lies in close proximity to a context-free language, a class characterized by its low rank in terms of formal complexity and expressive power~\cite{chomsky1956three}.
Perhaps more importantly, several commonly used evaluation test beds, including the Penn Treebank~\cite{klein2001parsing}, are in themselves context-free, muddying the territory between probing for acquired syntactic generalization and arbitrating pattern extraction.
As such, claims about the syntactic skills of language models should not be assumed to freely transfer between languages (and, in some cases, even datasets).

\begin{figure}[h]
    \centering
        {\smaller
        \begin{tikzpicture}[scale=0.7]
        \matrix [matrix of nodes](m){
        ... dat & Jan & Marie & de kinderen & ziet & leren & fietsen \\
        ... \textit{that} & \textit{Jan} & \textit{Marie} & \textit{the children} & \textit{see} & \textit{teach} & \textit{cycle}\\};
        \node[below of=m]{{\trans{`...that John sees Mary teach the kids to cycle'}}};
        \draw[-,color=blue] (m-1-4)..controls +(north:2) and +(north:2)..(m-1-7);
        \draw[-,color=green] (m-1-2)..controls +(north:2) and +(north:2)..(m-1-5);
        \draw[-,color=red] (m-1-3)..controls +(north:2) and +(north:2)..(m-1-6);
        \end{tikzpicture}
        }
    \caption{Illustration of crossing dependencies in Dutch.}
    \label{fig:cross_dep}
\end{figure}
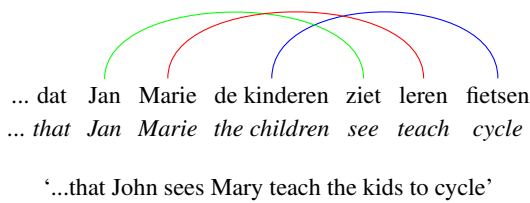

In this paper, we seek to evaluate BERT in the face of patterns that go beyond context-freeness.
We employ a \emph{mildly context-sensitive} grammar formalism to generate complex patterns that do not naturally occur in English.
We choose instead to experiment on Dutch, a language long-argued to be non-context free, due it its capacity for exhibiting an arbitrary number of \textit{cross-serial dependencies}.
In Dutch, cross-serial dependencies arise in sentences where verbs form clusters, causing their respective dependencies with their arguments to intersect when drawn on a plane: Figure~\ref{fig:cross_dep} portrays an adaptation of the example of~\citet{xdepsdutch}.%

\noindent To that end, we first identify two well-studied constructions in Dutch that commonly involve cross-serial dependencies: control verb nesting and verb raising. 
We produce an artificial but naturalistic dataset of annotated samples for each construction; each sample contains span annotations for the verb- and noun-phrases occurring within, as well as a mapping that associates each verb to its corresponding subject.
We then implement a probing model intended to select a verb's subject from a number of candidate phrases, train it on a gold-standard resource of Dutch, and employ it on our data.
Our experimental results convey a rapidly declining performance in the presence of discontinuous syntax, suggesting that the Dutch models investigated do not automatically learn to resolve the complex dependencies occurring in the language. To facilitate further research on the topic, our code is publicly available online.\footnote{\url{https://github.com/gijswijnholds/discontinuous-probing}.}

\section{Background}

\subsection{Context freeness of natural languages}
There has been a long debate, since the introduction of the Chomsky hierarchy \cite{chomsky1956three}, on whether all string patterns in natural language can be encompassed by the class of context-free grammars.
The dispute often makes a distinction between \textit{weak} and \textit{strong} context-freeness, whereby the question shifts between generating all strings or all constituent expressions of a language.
In Dutch specifically, patterns involving \emph{cross-serial dependencies} have been commonly brought up by linguists in arguing that at least fragments of Dutch are context-sensitive, in turn designating the language not strongly context-free~\cite{huybregts1984weak,pullum1982natural,xdepsdutch,shieber1985evidence}.

To capture such patterns without employing unnecessary computational expressiveness (and corresponding complexity), one can resort to the more pragmatic alternative of \textit{mildly} context-sensitive grammars~\cite{joshi1985tree}: systems that can capture certain types of crossing dependencies, while remaining computationally tractable.\footnote{
    Theoretical analyses of cross-serial dependencies can be found in various mildly context-sensitive frameworks, including CCG~\cite{steedman1985dependency}, Multimodal Typelogical Grammar~\cite{moortgat1999meaningful}, the Discontinuous Lambek Calculus~\cite{morrill2007dutch} and others~\cite{muskens2007separating,koopman2000verbal}.}

\subsection{Multiple Context-Free Grammars}
One of the more general classes of mildly context-sensitive systems are multiple context-free grammars (MCFGs), which essentially generalizes the notion of a context-free grammars to operations on \textit{tuples} of strings.
We defer the reader to~\citet{seki1991multiple} for a full definition and discussion of the properties of MCFGs.
Instead we provide a simplified, computationally-oriented description that is more in line with our purposes and implementation.
An $m$-multiple MCFG can be thought of as a tuple $\langle \mathcal{A}, \nts, d, \mathcal{C}, \rules, \nt{S}_0 \rangle$, 
where:
\begin{itemize}[noitemsep,topsep=0.3em]
    \item $\mathcal{A}$ is the terminal \textit{alphabet}
    \item \nts is a set of \textit{non-terminals} and ${d: \nts \to \mathbb{N}}$ a function from non-terminals to natural numbers; each non-terminal \nt{N} is encoding a tuple of strings of fixed arity $d(\nt{N})$ and the maximal arity of $\nts$ decides the grammar's multiplicity
    \item $\mathcal{C}$ is a mapping that associates each non-terminal \nt{N} to a (possibly empty) set of  elements from the $d(\nt{N})$-ary cartesian product ${(\mathcal{A}^*)}^{d(\nt{N})}$; put simply, the set of constants $\mathcal{C}_\nt{N}$ prescribes all the possible ways of initializing the non-terminal \nt{N}
    \item $\rules$ a set of \textit{rewriting rules}; rules are functions
    ${\nts \times \dots \times \nts \to \nts}$ that provide recipes on how to combine a number of non-terminals into a single non-terminal by rearranging and concetenating their contents; we will write:%
    \[
        \nt{C}(z_1,\dots z_k) \leftarrow \nt{A}{(x_1,\dots x_m}) ~\nt{B}{(y_1, \dots y_n})
    \]%
    to denote a rule that combines non-terminals \nt{A} and \nt{B} of arities $m$ and $n$ into a non-terminal \nt{C} of arity $k$, where each of the left-hand side coordinates $x_1,\dots  y_n$ is used exactly once in the right-hand side coordinates $z_1,\dots z_k$
    \item $\nt{S}_0$ the \textit{start} symbol, a distinguished element of $\nts$ satisfying $d(\nt{S}_0)= 1$
\end{itemize}%
The choice of MCFGs as our formal backbone comes due to their many advantages.
Being a subtle but powerful generalization of CFGs, MCFGs have a familiar presentation that makes them easy to reason about, while remaining computationally tractable~\cite{ljunglof-2012-practical,kallmeyer2010parsing}.
At the same time, they offer an appealing dissociation between abstract and surface syntax and lexical choice.
A derivation inspected purely on the level of rule type signatures takes the form of an abstract syntax tree that is reminiscent of a traditional CFG parse.
Normalizing an MCFG so as to disallow rules from freely inserting constant strings (i.e. wrapping all constants under a non-terminal) allows us to (i) trace back all substrings of the final yield to a single non-terminal and (ii) provide a clear computational interpretation that casts an MCFG as a linear type system, and its derivation as a functional program~\cite{de2003m}.

\section{Methodology}
\label{sec:methods}

\subsection{Linguistic background}
\label{subsec:lingback}

We focus on two patterns in Dutch: control verb nesting and verb raising.

\paragraph{Control Verb Nesting}
\label{par:control_cross}

Control verbs select a (referential) noun phrase and an infinitival complement which lacks an overt subject. This missing dependent (a so-called \textit{understood} subject) can be traced back to a higher level of the syntax tree, materialising as a dependent of the matrix clause; from a semantic standpoint, it is implicitly carried over to the subordinate clause by the control verb.
The choice of \textit{which} of the (possibly many) dependents is carried over is purely lexical, and essentially determined by the choice of verb~\cite{augustinus2015complement}\footnote{Some of the verbs that we select are optional clustering verbs, but we use them only in the control setting.}:

\ex.
    \label{ex:control}
    \ag.
    {de student}    {belooft}           {de docent}             {te} {studeren}   \\ 
	{\smaller the student} 	          {\smaller promises}   {\smaller the teacher}     {\smaller to} {\smaller study}      \\
	\trans{`the student promises the teacher to study'}
	\label{ex:control:belooft}
    \bg.
	{de docent}		vraagt 	{de student} 	{te} {studeren} \\
	{\smaller the teacher}	{\smaller asks}	{\smaller the student}	{\smaller to} {\smaller study} \\
	\trans{`the teacher asks the student to study'}
	\label{ex:control:vraagt}
    
The two sentences of example~\ref{ex:control} agree in their surface form, but differ in how the agent understood as `studying' is selected; in~\ref{ex:control:belooft} it is the main clause subject (`promise' being a \textit{subject control} verb), whereas in~\ref{ex:control:vraagt} it is the main clause object (`ask' being an \textit{object control} verb).

The basic constructions above can quickly become more nuanced in a variety of ways:

\ex.\label{ex:control:complicated}
	\ag.
	{de hond}	{vraagt}		{de student}		{de oefeningen}		{te} {eten}\\
	{\smaller the dog}       {\smaller asks}      {\smaller the student}       {\smaller the exercises}     {\smaller to} {\smaller eat}\\
	\trans{`the dog asks the student to eat the exercises'}%
	\label{ex:control:tv_inf}
	\bg.
	{de docent} {vraagt}    {de hond}   {de student} {de oefeningen} {te} {laten} {doen}\\
	{\smaller the teacher} {\smaller asks} {\smaller the dog} {\smaller the student} {\smaller the exercises} {\smaller to} {\smaller let} {\smaller do}\\
    \trans{`the teacher asks the dog to let the student do the exercises'}%
    \label{ex:control:aux_tv}
    \cg.
    {de docent} {vraagt}    {de hond}   {de student} {te} {beloven} {de oefeningen}  {niet} {te} {eten}\\
    {\smaller the teacher}  {\smaller asks} {\smaller the dog} {\smaller the student} {\smaller to}
    {\smaller promise}  {\smaller the exercises} {\smaller not} {\smaller to} {\smaller eat}\\
    \trans{`the teacher asks the dog to promise the student not to eat the exercises'}
    \label{ex:control:rec}
    
To begin with, if the head of the subordinate clause is a transitive infinitive, its object is positioned immediately after the main clause; this has the effect of creating a sequence of noun phrases that precede the verbal complement~\ref{ex:control:tv_inf}.
Further, in the case of the infinitive being a causative which selects for another infinitive, subject selection is preserved for the former, but flipped for the latter~\ref{ex:control:aux_tv}.

Finally, things get interesting when realizing that the above patterns can recurse, as a verbal complement may act as the object of another verbal complement~\ref{ex:control:rec}.

The nesting of control verbs makes for a challenging probing task, as the dependency between a verb and its subject may span multiple depths of the syntax tree, while at the same time requiring subtle lexical distinctions to resolve correctly.

\begin{figure*}[h!]
    \begin{subfigure}{1\textwidth}
    \begin{center}
        \begin{align}
            \mcfgrule{\nt{S}(xyzu_1u_2)}{
                \nt{NP}(x) ~ 
                \nt{TV}(y) ~
                \nt{NP}(z) ~
                \nt{VC}(u_1,u_2)}\tag{\ensuremath{A_1}}\label{gr:control:1}\\
            \mcfgrule{\nt{S}(xyzuw_1vw_2)}{
                \nt{NP}(x) ~
                \nt{TV}(y) ~
                \nt{NP}(z) ~
                \nt{NP}(u) ~
                \nt{CV}(v) ~
                \nt{VC}(w_1,w_2)}\tag{\ensuremath{A_2}}\label{gr:control:2}\\
            \mcfgrule{\nt{VC}(x,y)}{
                \nt{TE}(x) ~
                \nt{INF}_{iv}(y)}\tag{\ensuremath{A_3}}\label{gr:control:3}\\
            \mcfgrule{\nt{VC}(zx, y)}{
                \nt{TE}(x) ~
                \nt{INF}_{tv}(y) ~
                \nt{NP}(z)}\tag{\ensuremath{A_4}}\label{gr:control:4}\\
            \mcfgrule{\nt{VC}(xy,zu_0u_1)}{
                \nt{NP}(x)~
                \nt{TE}(y)~
                \nt{INF}_{c}(z)~
                \nt{VC}(u_0,u_1)}\tag{\ensuremath{A_5}}\label{gr:control:5}\\
            \mcfgrule{\nt{VC}(xyu,zv_1v_2)}{
                \nt{NP}(x)~
                \nt{TE}(y)~
                \nt{INF}_c(z)~
                \nt{CV}(u)~
                \nt{VC}(v_1,v_2)}\tag{\ensuremath{A_6}}\label{gr:control:6}\\
            \midrule
            \mcfgrule{\nt{S}(xyzvu_1u_2)}{
                \nt{NP}(x)~
                \nt{TV}(y)~
                \nt{NP}(z)~
                \nt{VC}(u_1,u_2)~
                \nt{ADV}(v)}\tag{$A_1^{m}$}\label{gr:control:1a}\\
            \mcfgrule{\nt{S}(vyxzu_1u_2)}{
                \nt{NP}(x)~
                \nt{TV}(y)~
                \nt{NP}(z)~
                \nt{VC}(u_1,u_2)
                ~\nt{ADV}(v)}\tag{$A_1^{i}$}\label{gr:control:1b}\\
            & \quad\quad\vdots\notag
        \end{align}
    \end{center}\vspace{-\baselineskip}
    \caption{2-MCFG for control verbs.}
    \label{fig:grammar_control}
    \end{subfigure}
    \begin{subfigure}{1\textwidth}
    \begin{center}
        \begin{align}
            \mcfgrule{\nt{S}(xy_1y_2)}{
                \nt{PREF}(x) ~ 
                \nt{SUB}(y_1,y_2)}\tag{\ensuremath{B_1}}\label{gr:cluster:1}\\
            \mcfgrule{\nt{SUB}(x, y)}{
                \nt{NP}(x) ~
                \nt{INF}_{iv}(y)}\tag{\ensuremath{B_2}}\label{gr:cluster:2}\\
            \mcfgrule{\nt{SUB}(xy, z)}{
                \nt{NP}(x) ~
                \nt{NP}(y) ~
                \nt{INF}_{tv}(z)}\tag{\ensuremath{B_3}}\label{gr:cluster:3}\\
            \mcfgrule{\nt{SUB}(xz, yu)}{
                \nt{NP}(x) ~
                \nt{RV}(y) ~
                \nt{SUB}(z, u)}\tag{\ensuremath{B_4}}\label{gr:cluster:4}
        \end{align}
    \end{center}
    \caption{2-MCFG for verb raising.}
    \label{fig:grammar_cluster}
    \end{subfigure}
    \caption{2-MCFGs capturing the phenomena of Section~\ref{subsec:lingback}.}
    \label{fig:grammars}
\end{figure*}

\paragraph{Verb Raising}
\label{par:raising}
Dutch verb raising is the phenomenon whereby the head of an infinitival complement attaches to the verb governing it, creating a cluster in the process~\cite{evers1976transformational}.
Verbs allowing this construction select for bare complements (i.e. do not require the complementizer \textit{te}).
Unlike the previous case, the verbal complement does now contain a material subject; the complication is this time due to each nested verbal complement adding yet another set of crossing dependencies.

\ex.\label{ex:raising}
	\ag.
	{de docent} {ziet} {de student} {de hond}	{de oefeningen}	 {leren} {eten} \\
	{\smaller the teacher} {\smaller sees} {\smaller the student} {\smaller the dog}       {\smaller the exercises}  {\smaller teach}    {\smaller eat}\\
	\trans{`the teacher sees the student teach the dog to eat the exercises'}%
    \label{ex:raising:two}
    \bg.
        {de docent} 
        {ziet} 
        {de hond} 
        {de student} 
        {de eend} 
        {de oefeningen} 
        {helpen} 
        {leren} 
        {eten}\\
    {\smaller the teacher}
    {\smaller sees}
    {\smaller the dog}
    {\smaller the student}
    {\smaller the duck}
    {\smaller the exercises}
    {\smaller help}
    {\smaller teach}
    {\smaller eat}\\
	\trans{`the teacher sees the dog help the student teach the duck to eat the exercises'}
    \label{ex:raising:three}
    
By construction, the verb raising grammar isolates the problem of resolving verb-subject dependencies in a purely syntactic setting, as no lexical variation will change the choice of dependent for a given verb.
As such, it allows us to probe for a model's potential at syntactic generalization that does no longer rely on lexical cues.

\subsection{Data generation}
\label{subsec:data_gen}

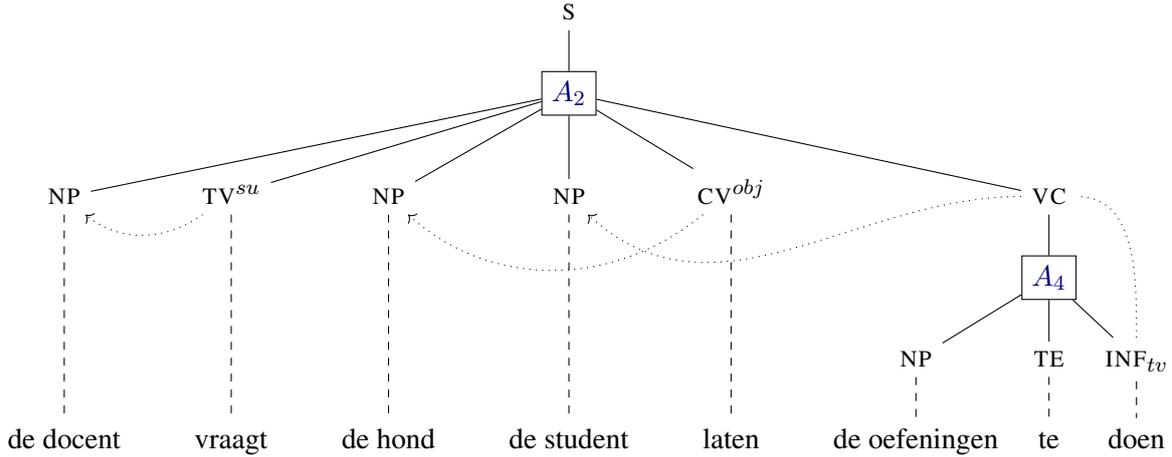
\begin{figure*}
    \centering
    \begin{forest}
    rule/.style={draw=black},
    nt/.style={}
    [\nt{S},
        [{\ref{gr:control:2}},calign=child,calign child=4,l sep=20pt,s sep=20pt,rule
            [\nt{NP},name={s1},nt
                [{de docent},tier=word,edge=dashed]
            ]
            [$\nt{TV}^{su}$,name={v1},nt
                [{vraagt},tier=word,edge=dashed]
            ]
            [\nt{NP},name={s2},nt
                [{de hond},tier=word,edge=dashed]
            ]
            [\nt{NP},name={s3},nt
                [{de student},tier=word,edge=dashed]
            ]
            [$\nt{CV}^{obj}$,name={v2},nt
                [{laten},tier=word,edge=dashed]
            ]
            [\nt{VC},name={v3},nt,l sep=1pt
                [\ref{gr:control:4},calign=child,calign child=2,rule,l sep=1pt
                    [\nt{NP},nt
                        [{de oefeningen},tier=word,edge=dashed]
                    ]
                    [\nt{TE},nt
                        [{te},edge=dashed]
                    ]
                    [\vphantom{TE}{$\nt{INF}_{tv}$},name={v4},nt
                        [{doen},tier=word,edge=dashed]
                    ]
                ]
            ]
        ]
    ]
    \draw[->,dotted] (v1) to[out=south west,in=south east]  (s1);
    \draw[->,dotted] (v2) to[out=south west,in=south east]  (s2);
    \draw[->,dotted] (v3) to[out=west,in=south east]  (s3);
    \draw[dotted] (v4) to[out=north, in=east]            (v3);
    \end{forest}
    \caption{Generation tree for example~\ref{ex:control:aux_tv}. 
    Boxed nodes correspond to rule applications. 
    Non-terminal superscripts denote verbal subtype (subject- or object control). 
    Dashed lines assign lexical constants to non-terminals. 
    Dotted lines demonstrate how verbs select for their subjects: $\nt{TV}^{su}$ and $\nt{CV}^{obj}$ both find their subjects at the same depth of the tree, but the presence of the latter signifies that the main clause object will be propagated to the verbal complement, to be there selected by $\nt{INF}_{tv}$.
    Note that the tree presented should not be confused for a constituency parse -- a more fitting paradigm would be an abstract syntax tree, that prescribes the program
    $\ref{gr:control:2}\left(\nt{NP}(\text{\smaller de docent}), \nt{TV}(\text{\smaller vraagt}), \nt{NP}(\text{\smaller de hond}), \nt{NP}(\text{\smaller de student}), \nt{CV}(\text{\smaller laten}), \ref{gr:control:4}\left(\nt{NP}(\text{\smaller de oefeningen}), \nt{TE}(\text{\smaller te}), \nt{INF}_{tv}(\text{\smaller doen}) \right) \right) \mapsto \ref{ex:control:aux_tv} $.}
    \label{fig:example_tree}
\end{figure*}

For our data generation needs, we design a custom implementation of an MCFG enriched with two added functionalities.
First, we define two sets ${\vs,~ \ns \subset \nts}$ that specify which non-terminals correspond to verb- and noun phrases respectively.
Every occurrence of a marked non-terminal indicates a unique phrase in the final yield, which we can trace by traversing the derivation tree.
This, in turn, gives us the possibility of assigning one or more labels to the constituent substrings that make up a sentence, according to which phrase(s) they were part of, even in the case of discontinuous and/or overlapping substrings.
Additionally, we decorate MCFG rules with subject inheritance schemes.
In the simplest case, a scheme may directly specify the subject noun of a verb, if the non-terminals of both occur on the same rule, i.e. they inhabit the same depth of the generation tree.
Alternatively, when the two occur at different depths, a scheme may defer the decision by propagating verb indices down through non-nominal constituents that will contain the matching subject, but at an arbitrary nesting depth (see Figure~\ref{fig:example_tree} for an example).
Lexical constants for primitive categories are populated by means of an automatically compiled but manually verified lexicon.
    
\subsection{Grammars}
\label{subsec:grammars}
% TODO
We use the above framework to instantiate distinct grammars for both syntactic phenomena of interest.
Note that the grammars are not purposed for the construction of exhaustive or accurate analyses of the phrase structures considered, but rather for the controlled generation and annotation of suitable samples.

\paragraph{Control Verb grammar} 
Our first grammar, given in Figure~\ref{fig:grammar_control}, models control verb nesting.
The grammar accounts for the mobility of verbal complements by encoding them as non-terminals of multiplicity 2, making the grammar a 2-MCFG.
We have two constructors for sentences that combine two noun phrases and a transitive verb with a verbal complement (\ref{gr:control:1}), optionally under the context of a causative verb and its direct object (\ref{gr:control:2}).
In the base case, verbal complements are constructed with \textit{te} and either an intransitive infinitive (\ref{gr:control:3}) or a transitive infinitive and its object (\ref{gr:control:4}).
In the inductive case, a verbal complement can contain a control verb in infinitival form together with a noun phrase and another verbal complement, either alone or with a causative (\ref{gr:control:5} and~\ref{gr:control:6}).
To increase the variance of generated samples, we also consider two variations for each of the first two rules that incorporate adverbial modifiers: one where the adverb is inserted after the verb (\ref{gr:control:1a}) and, more interestingly, one where the adverb is inserted before the verb (\ref{gr:control:1b}); Dutch being a V2 language, this has the effect of inverting the position of the verb and subject of the main clause.

We set $\vs:=\left\{ \nt{TV}, \nt{MV}, \nt{INF}_x \right\}$ and ${\ns:=\left\{\nt{NP}\right\}}$.
We divide each of \nt{TV} \nt{MV} and $\nt{INF}_c$ into two subtypes, specifying whether they are subject- or object-selecting; each subtype has a distinct set of lexical entries.
Finally we decorate each rule with subject propagation schemes, dependent on the subtypes of the participating verbal non-terminals; rather than explicitly enumerate these schemes here, we provide a visual example in Figure~\ref{fig:example_tree}.

\paragraph{Verb Raising grammar} For the second grammar we can do with just four rules (Figure \ref{fig:grammar_cluster}).
The grammar is centered around a single non-terminal of multiplicity 2 that encodes subordinate clauses.
In the base case, such a clause can be constructed with the aid of either a noun-phrase and an intransitive infinitive~(\ref{gr:cluster:2}), or two noun phrases and a transitive~(\ref{gr:cluster:3}).
In the inductive case, a subordinate clause is embedded within a broader subordinate clause, where it occupies the object position of a raising verb~(\ref{gr:cluster:4}).
Finally, a sentence is generated by joining a subordinate clause to a matrix clause missing its verbal complement -- we avoid deconstructing the matrix clause and denote it as a fixed prefix string~(\ref{gr:cluster:1}).
We set $\vs:=\left\{ \nt{INF}_{iv}, \nt{INF}_{tv}, \nt{RV} \right\}$ and ${\ns:=\left\{\nt{NP}\right\}}$. 
Unlike in the case of control verb nesting, there is no subject inheritance necessary; rules~\ref{gr:cluster:2},~\ref{gr:cluster:3}, and~\ref{gr:cluster:4} all add a verb and their subject simultaneously.

\subsection{Probing Model}
Our probing model first aggregates the contextualized token representations for each verb- and noun-phrase, before computing a verb-to-noun cross-attention matrix.

The aggregation process is essentially an attentive pooling over (two types of) variably sized, potentially overlapping clusters~\cite{li2015gated}.
We start by representing each distinct verb- and noun-phrase as a binary mask over the tokenized input sentence; each sentence is then associated with a variable number of both types of masks.
Using a pair of learned projections, we map the BERT-contextualized token representations into scalar values denoting attention scores.
For each phrase, attention weights for participating (potentially discontinuous) tokens are computed by softmaxing their corresponding attention scores; summing the attention-weighted BERT representations yields a single vector for each phrase.
We use the implementation of~\citet{Fey/Lenssen/2019} to efficiently compute batch-wide representations leveraging the sparsity of the phrasal masks.

The pair-wise agreement between verb and noun representations is computed using standard dot-product attention, restricted to pairs occurring in the same sentence via dynamic masking.
Prior to computing this attention matrix, we map the verb and noun representations to a lower dimension using another pair of learned projections; this serves to add a hint of expressive capacity to the probe, while also reducing the memory footprint of the matrix multiplication. 
Finally, softmaxing the attention weights over the noun-dimension allows us to retrieve a trainable subject selection for each occurrence of a verb.

\section{Experiments \& Results}
\subsection{Experimental setup}

The experiments with our grammars consist of several parts. 
We first carry out an automatic filtering and annotation process on a gold standard corpus to gather a collection of suitable sentences, with which we train our probe on a natural, ``real-world'' dataset.
To obtain our datasets, we start by fixing a maximal recursion depth for each grammar, and exhaustively generate the corresponding sets of derivable abstract trees.
We then semi-automatically assemble a lexicon, with which we populate the various primitive categories employed by our grammars.
From each tree, we obtain a set number of unique sentences by randomly sampling the constants behind leaf non-terminals with a preset seed.
Finally, we apply the trained probe on the artificial samples and measure its performance across various generation parameters.

\paragraph{Probe Training}
An inevitable downside of using a rule-based system for generation purposes is low variance in several aspects of the output data.
In our case, the limited number of rules employed, in combination with their relative simplicity, would mean a fair amount of repeating patterns that are easy to decipher and memoize.
Albeit an advantage for interpretability and analysis purposes, this can potentially backfire if we are to use our grammars' yield for training: one can assume that BERT's contextualization preserves, at least in part, the relative position information contained within its input, thus providing the probe with a workaround (or confound) to the actual problem.
To avoid overfitting, we consequently choose to train the probe on an external data source derived from Lassy-Small, the gold standard corpus of written Dutch~\cite{van2013large}.
Lassy makes for an excellent data source for our task, as it provides analyses in the form of graphs, rather than trees, so as to explicitly account for several non-local phenomena (crucially, this includes the semantic subjects of verbal  complements).
We traverse the Lassy graphs to annotate noun phrases (all leaf nodes that descent from a noun phrase or are otherwise marked as a noun or pronoun) and verbs of interest (phrasal heads within a dependency frame that contains a subject previously identified as a noun phrase).
From the 65\,000 samples of Lassy, we extract about 12\,000 that contain at least two \textit{distinct} subjects without exceeding a word length of 30.
We split the latter into two mutually exclusive sets of 10\,000 and 2\,000 samples: we train with the first and use the second for model selection.

We experiment with two Dutch language models: BERTje~\cite{de2019bertje} and RobBERT~\cite{delobelle2020robbert}, based on BERT and RoBERTa~\cite{liu2019roberta} respectively. BERTje and RobBERT have shown to perform highly on a variety of Dutch NLP tasks, such as Named Entity Recognition \cite{sang2003introduction}, Sentiment Analysis \cite{vanderburgh2019merits},  and Natural Language Inference \cite{wijnholds-moortgat-2021-sick}.
For each model, we train 3 probes that differ in their initialization seeds, using AdamW~\cite{loshchilov2018decoupled} with a learning rate of $10^{-4}$, a batch size of 32 and a dropout rate of 15\%, applied at BERT's output.
We perform model selection using accuracy over the validation set as our metric, measured over individual verb predictions; validation accuracy converges after ca. 80 training epochs.

\begin{table*}
    \renewcommand{\arraystretch}{0.9}
    \begin{subtable}{1\textwidth}
    \begin{tabularx}{0.99\textwidth}{@{}L@{\qquad}CCCC@{\qquad}CCC@{\qquad}CCCCCCCC@{}}
        % # nouns
        & \multicolumn{4}{c}{\textbf{\smaller \# Nouns}} 
        % depth
        & \multicolumn{3}{c}{\textbf{\smaller Tree Depth}} 
        % rule s
        & \multicolumn{6}{c}{\textbf{\smaller Rule}} \\
        \textbf{\smaller Model}
        % # nouns
        & {\smaller 2} & {\smaller 3} & {\smaller 4} & {\smaller 5} 
        % depth
        & {\smaller 2} & {\smaller 3} & {\smaller 4} 
        % rules
        & {\smaller \hyperref[gr:control:1]{$A_1^X$}} & {\smaller \hyperref[gr:control:2]{$A_2^X$}} & {\smaller \ref{gr:control:3}} & {\smaller \ref{gr:control:4}} & {\smaller \ref{gr:control:5}} & {\smaller \ref{gr:control:6}} \\
    \toprule
    {\smaller BERTje}
        % # nouns
        & \nr{81.1}{2.6} & \nr{58.8}{2.6} & \nr{50.5}{0.5} & \nr{42.9}{1}
        % depth
        & \nr{61.8}{1} & \nr{52.7}{0.3} & \nr{46.8}{0.7}
        % rules
        & \nr{100}{0} & \nr{67}{2.1} & \nr{43.1}{1} & \nr{34.6}{3} & \nr{36.1}{1.3} & \nr{27.1}{1.2} \\
    {\smaller RobBERT}
        % # nouns
        & \nr{73}{1} & \nr{52.8}{0.6} & \nr{42.4}{0.5} & \nr{35.9}{0.7}
        % depth
        & \nr{58.3}{1.3} 
        & \nr{47.2}{0.1} 
        & \nr{38.8}{0.6} 
        % rules
        & \nr{93.6}{0.7} & \nr{58.1}{1.4} & \nr{41.2}{0.5} & \nr{19.6}{0.7} & \nr{21}{1} & \nr{17}{0.8}
    \end{tabularx}
    \label{tab:numbers:control}
    \caption{Control Verb Grammar}
    \end{subtable}
    \begin{subtable}{1\textwidth}
    \begin{tabularx}{0.99\textwidth}{@{}L@{\qquad}CCCC@{\qquad}CCCCC@{\qquad}CCC@{}}
        % # nouns
        & \multicolumn{4}{c}{\textbf{\smaller \# Nouns}} 
        % depth
        & \multicolumn{5}{c}{\textbf{\smaller Tree Depth}} 
        % rule s
        & \multicolumn{3}{c}{\textbf{\smaller Rule}} \\
        \textbf{\smaller Model}
        % # nouns
        & {\smaller 2} & {\smaller 3} & {\smaller 4} & {\smaller 5} 
        % depth
        & {\smaller 2} & {\smaller 3} & {\smaller 4} & {\smaller 5} & {\smaller 6}
        % rules
        & {\smaller \ref{gr:cluster:2}} & {\smaller \ref{gr:cluster:3}} & {\smaller \ref{gr:cluster:4}} \\
    \toprule
    {\smaller BERTje}
        % # nouns
        & \nr{75.6}{2.4} & \nr{52.4}{2.4} & \nr{33.5}{1.9} & \nr{25.5}{1.1}
        % depth
        & \nr{92.2}{1.6} & \nr{66.4}{2.8} & \nr{40.5}{0.9} & \nr{29}{2.4} & \nr{23}{1.9}
        % rules
        & \nr{53.4}{3.5} & \nr{53.8}{1.8} & \nr{36.7}{0.8} \\ 
    {\smaller RobBERT}
        % # nouns
        & \nr{46.3}{2.9} & \nr{37.2}{2.4} & \nr{24.5}{1} & \nr{11.4}{1.5}
        % depth
        & \nr{65.6}{2} & \nr{36.9}{3.7} & \nr{33.6}{1.1} & \nr{19.4}{1} & \nr{12.6}{1.4}
        % rules
        & \nr{89.1}{1.4} & \nr{24.3}{1.8} & \nr{12.9}{1.9} \\ 
    \end{tabularx}
    \label{tab:numbers:cluster}
    \caption{Verb Raising Grammar}
    \end{subtable}
    \caption{Seed-averaged accuracy scores for the two grammars of Section~\ref{subsec:grammars}, grouped by various parameters. The $X$ superscript denotes inclusion and aggregation of the adverbial modifier variants for the corresponding rules.}
    \label{tab:numbers}
\end{table*}

\paragraph{Controlled data generation}
Despite remaining grammatical, sentences start looking odd and unnatural when allowing recursion to arbitrary depth -- we impose an upper limit that leads to complex but still human-parsable data: 4 and 6 for the verbal control and raising grammars, respectively.
To cast the generated trees into sentences, we populate primitive categories (that is, categories that can be instantiated lexically rather than -- or in addition to --  by rule) with sets of semi-automatically assembled constants.
For simplicity, we consider only the case of verbs accepting a person as their indirect object; we filter 40 such verbs from a larger collection of ditransitives crawled from Lassy, and manually gather 30 temporal, locative and manner adverbs that can modify them.
All verbs are drawn from Lassy~\cite{van2013large} and the lists of~\citet{augustinus2015complement}: we gather 9 subject- and 33 object-control verbs, 2 causatives and 7 raising verbs.
A comprehensive set of around 10\,000 gendered nouns (the ones that have \textit{de} as their article) is finally obtained from the \emph{Algemeen Nederlands Woordenboek}.\footnote{\url{https://anw.ivdnt.org}}
In the verb raising grammar, we trigger subordination by prefixing generated expressions with the string \textit{Iemand ziet} (\trans{`somebody sees'}).

From each generated abstract tree, we obtain 10 syntactically identical sentences that vary only in their meaning by performing controlled sampling over the lexicon; the very large product space of constants guarantees sample uniqueness.
This parameterization means we can inspect and group samples on the basis of either their surface form or their underlying tree, a property that will come when analyzing model performance.
To ensure naturality and consistency in the model's input, we capitalize and punctuate generated sentences in a final post-processing step.

\subsection{Results}
The trained probes are tested on our generated data, yielding a prediction for every verb occurrence.
For each model, we report the seed-averaged accuracy on each experiment in Table~\ref{tab:numbers:general}: test performance is substantially lower than in the validation benchmarks.

\begin{table}[h]
    \centering
    \begin{tabularx}{0.45\textwidth}{@{}L@{\qquad}CCC@{}}
        \textbf{\smaller Model} & \textbf{\smaller Lassy} & \textbf{\smaller Control} & \textbf{\smaller Raising}\\ 
        \toprule
        {\smaller BERTje}   & \nr{97.6}{}   & \nr{48}{}     & \nr{43.1}{}  \\
        {\smaller RobBERT}  & \nr{92.5}{}   & \nr{40.6}{}   & \nr{29.2}{}
    \end{tabularx}
    \caption{Model accuracy on the validation data (Lassy) versus the test data (Control, Raising).}
    \label{tab:numbers:general}
\end{table}

To facilitate analysis, we group predictions according to their context, namely (i) number of noun phrases in the sentence (classification targets), (ii) maximal depth of the underlying abstract syntax tree and (iii) production rule, and aggregate them into accuracy scores, presented in Table~\ref{tab:numbers}.
This breakdown suggests that model performance remains passable for the easier portion of the dataset, but degrades quickly as the difficulty of the task increases; models have a harder time associating a verb to its subject as sentences get longer and more complicated.
The over-representation of harder-samples due to the dominance of deeper abstract syntax trees then serves to explain the striking performance decline.

\paragraph{Control Verbs}
Focusing on the control grammar first, we remark that both models consistently score above the random baseline (i.e. 1 divided by the number of classification targets), seemingly indicating that some notion of semantic comprehension perseveres in the presence of control verb nestings.
Grouping scores by rule is revealing: the main clause subject is (almost) always correctly detected, regardless of nestedness of the co-occurring complement and unperturbed by the presence of word-order variations due to modifiers (\hyperref[gr:control:1]{$A_1^X$}).
Verbal complements and causatives, on the other hand, are more often than not incorrectly analyzed, even in the simplest cases of a bare infinitive in isolation (\ref{gr:control:3}), or a causative occurring at the topmost branch of the tree (\hyperref[gr:control:2]{$A_2^X$}).

\begin{table}[h]
\centering
    \begin{tabularx}{0.43\textwidth}{@{}Lccc@{}}
        & \multicolumn{2}{c}{\textbf{\smaller Control Scope}} & \multirow{2}{*}{\textbf{\smaller Consistency}} \\ 
        \textbf{\smaller Model} & {\smaller subject} & {\smaller object} \\
        \toprule
        {\smaller BERTje}  & \nr{34.4}{2.2} & \nr{36.1}{0.8} & \nr{68.4}{21.9} \\
        {\smaller RobBERT} & \nr{18.7}{0.7} & \nr{37.8}{5.3} & \nr{63}{22.2}
    \end{tabularx}
    \caption{Metrics specific to the Control Verb grammar.}
    \label{tab:ctrl}
\end{table}

To procure an explanation for this discrepancy, we start by measuring accuracy in verbs occurring under subject- and object control scopes separately.
The remarkably low results hint that models struggle with both kinds of control, while indicating the presence of an implicit bias slightly favouring the more common object control reading (especially so in the case of RobBERT).
Next, we investigate whether the low performance is due to models simply misreading certain constructions, assigning subjecthood to the (same) wrong noun phrase.
To quantify how consistent the models are, we gather all predictions occurring in the same context (i.e. same part of the same tree under the same scope, object or subject) and varying only in terms of lexical realization.
The consistency of a model in a specific context is calculated as the frequency of the most common prediction (correct or otherwise); the model's overall consistency is then the average consistency over all contexts.
Models generally fail at producing the same prediction given the same syntactic template, instead being susceptible to distraction from word variations.

\paragraph{Verb Raising}
The story is no different when it comes to the second grammar: both models fail to draw close to their validation benchmarks.
Surprisingly, RobBERT's metrics lie below the random baseline, positing that it encodes a wrong syntactic structure in verb cluster formations, rather than simply not acquiring the correct one.
The disproportionately high accuracy of rule~\ref{gr:cluster:2} readily provides an explanation: the noun-phrase directly preceding an infinitive is assumed to be its subject.
BERTje, on the other hand, is more trustworthy, maintaining comparable performance in both intransitive~(\ref{gr:cluster:2}) and transitive~(\ref{gr:cluster:3}) infinitival phrases. 
The degradation associated with deeper trees coincides with the drop in performance for the recursive rule~\ref{gr:cluster:4}.

\subsection{One-Shot Learning}
Given the purported inadequacy of both models at correctly or consistently predicting subjecthood in our datasets' cross-serial constructions, we resort to one final experiment that serves as a sanity check for the quality of our data.
Using a different lexical sampling seed, we generate a single sentence from each abstract syntax tree, resulting in datasets of 307 and 30 samples for the control verb and verb raising grammars, respectively.
These compact datasets are then used for fine-tuning the two models (combined with probes) in a one-shot learning fashion; after a few epochs of training, we test the resulting models on the corresponding original datasets.

\begin{table}[h]
    \centering
    \begin{tabularx}{0.33\textwidth}{@{}L@{\qquad}CC@{}}
        \textbf{\smaller Model} & \textbf{\smaller Control} & \textbf{\smaller Raising}\\ 
        \toprule
        {\smaller BERTje} & \nr{92.4}{--} & \nr{68.5}{--} \\
        {\smaller RobBERT} & \nr{61.6}{--} & \nr{36.7}{--}
    \end{tabularx}
    \caption{Model results for the one-shot setup.}
    \label{tab:finetuning}
\end{table}

The results, presented in Table~\ref{tab:finetuning}, show that minimal supervision does improve model performance, indicating that the learned parameter updates generalize beyond the lexical choices of the fine-tuning data, thereby verifying the generation pipeline's internal consistency.
Improvement is lower in the case of the verb raising grammar; we posit that the task is harder to acquire due to its predominantly syntactic nature but also the smaller number of training samples.

\section{Conclusion}
We implemented a test suite based on multiple context-free grammars to generate a large collection of sentences containing complicated syntactic phenomena specific to Dutch.
We trained a probing model on extracting verb-to-subject pairings from the contextualized representations of state-of-the-art pretrained Dutch language models using an external resource of generic text accompanied by gold standard annotations.
We then tested the probe on our generated data, and found it to perform substantially below its own validation benchmarks.
After conducting extensive analysis aimed at identifying the source of this discrepancy, we showed that the probe's predictions are inconsistent and its accuracy quickly diminishes as the complexity of the syntactic patterns increases.
Based on the above, we conclude that neither of the BERT models investigated has learned to internalize syntactic and semantic subjecthood in nested constructions involving cross-serial dependencies.
Our findings serve as empirical evidence hinting at unsupervised language models having difficulty in the automatic acquisition of discontinuous syntactic patterns.

We leave several directions open for future work.
To begin with, one could mirror the patterns analyzed to other languages and compare model performance cross-linguistically, juxtaposed by the corresponding grammar complexity.
Alternatively, one could render more elaborate grammars intended to capture other syntactic or semantic phenomena of interest.
Finally, it is worth investigating the extent to which the ``real-world'' validation samples incorrectly classified are exemplars of the types of discontinuity captured by our grammars.

\section*{Acknowledgments}

The authors are grateful to the anonymous reviewers for their feedback. The authors moreover thank Michael Moortgat for review of the paper before submission, and for valuable discussions. Both authors acknowledge support from the Dutch Research Council (NWO) under the scope of the project “A composition calculus for vector-based semantic modelling with a localization for Dutch” (360-89-070).

% Entries for the entire Anthology, followed by custom entries
\bibliography{anthology,custom}

\begin{thebibliography}{34}
\expandafter\ifx\csname natexlab\endcsname\relax\def\natexlab#1{#1}\fi

\bibitem[{Augustinus(2015)}]{augustinus2015complement}
Liesbeth Augustinus. 2015.
\newblock \href {https://www.lotpublications.nl/Documents/413_fulltext.pdf}
  {\emph{Complement raising and cluster formation in {Dutch}}}.
\newblock Netherlands Graduate School of Linguistics.

\bibitem[{Bresnan et~al.(1982)Bresnan, Kaplan, Peters, and Zaenen}]{xdepsdutch}
Joan Bresnan, Ronald~M. Kaplan, Stanley Peters, and Annie Zaenen. 1982.
\newblock \href {http://www.jstor.org/stable/4178298} {Cross-serial
  dependencies in {Dutch}}.
\newblock \emph{Linguistic Inquiry}, 13(4):613--635.

\bibitem[{Chen et~al.(2021)Chen, Fu, Xu, Xie, Tan, Chen, and
  Jing}]{chen2021probing}
Boli Chen, Yao Fu, Guangwei Xu, Pengjun Xie, Chuanqi Tan, Mosha Chen, and
  Liping Jing. 2021.
\newblock \href {https://openreview.net/forum?id=17VnwXYZyhH} {Probing
  {\{}bert{\}} in hyperbolic spaces}.
\newblock In \emph{International Conference on Learning Representations}.

\bibitem[{Chomsky(1956)}]{chomsky1956three}
Noam Chomsky. 1956.
\newblock \href {https://www.its.caltech.edu/~matilde/Chomsky3Models.pdf}
  {Three models for the description of language}.
\newblock \emph{IRE Transactions on information theory}, 2(3):113--124.

\bibitem[{De~Groote and Pogodalla(2003)}]{de2003m}
Philippe De~Groote and Sylvain Pogodalla. 2003.
\newblock \href
  {https://hal.inria.fr/file/index/docid/107690/filename/A03-R-243.pdf}
  {m-linear context-free rewriting systems as abstract categorial grammars}.
\newblock In \emph{Proceedings of Eighth Meeting on Mathematics of Language
  (MOL 8)}, pages 71--80.

\bibitem[{de~Vries et~al.(2019)de~Vries, van Cranenburgh, Bisazza, Caselli, van
  Noord, and Nissim}]{de2019bertje}
Wietse de~Vries, Andreas van Cranenburgh, Arianna Bisazza, Tommaso Caselli,
  Gertjan van Noord, and Malvina Nissim. 2019.
\newblock \href {https://arxiv.org/pdf/1912.09582.pdf} {{BERT}je: A {Dutch}
  {BERT} model}.
\newblock \emph{arXiv preprint arXiv:1912.09582}.

\bibitem[{Delobelle et~al.(2020)Delobelle, Winters, and
  Berendt}]{delobelle2020robbert}
Pieter Delobelle, Thomas Winters, and Bettina Berendt. 2020.
\newblock \href {https://aclanthology.org/2020.findings-emnlp.292.pdf}
  {Rob{BERT}: a {Dutch} ro{BERT}a-based language model}.
\newblock In \emph{Proceedings of the 2020 Conference on Empirical Methods in
  Natural Language Processing: Findings}, pages 3255--3265.

\bibitem[{Devlin et~al.(2019)Devlin, Chang, Lee, and
  Toutanova}]{devlin-etal-2019-bert}
Jacob Devlin, Ming-Wei Chang, Kenton Lee, and Kristina Toutanova. 2019.
\newblock \href {https://doi.org/10.18653/v1/N19-1423} {{BERT}: Pre-training of
  deep bidirectional transformers for language understanding}.
\newblock In \emph{Proceedings of the 2019 Conference of the North {A}merican
  Chapter of the Association for Computational Linguistics: Human Language
  Technologies, Volume 1 (Long and Short Papers)}, pages 4171--4186,
  Minneapolis, Minnesota. Association for Computational Linguistics.

\bibitem[{Evers et~al.(1976)}]{evers1976transformational}
Arnold Evers et~al. 1976.
\newblock The transformational cycle in {D}utch and {G}erman.
\newblock \emph{Nieuwe (De) Taalgids}, 69(2):156--160.

\bibitem[{Fey and Lenssen(2019)}]{Fey/Lenssen/2019}
Matthias Fey and Jan~E. Lenssen. 2019.
\newblock \href {https://arxiv.org/pdf/1903.02428.pdf} {Fast graph
  representation learning with {PyTorch Geometric}}.
\newblock In \emph{ICLR Workshop on Representation Learning on Graphs and
  Manifolds}.

\bibitem[{Hewitt and Manning(2019)}]{hewitt-manning-2019-structural}
John Hewitt and Christopher~D. Manning. 2019.
\newblock \href {https://doi.org/10.18653/v1/N19-1419} {{A} structural probe
  for finding syntax in word representations}.
\newblock In \emph{Proceedings of the 2019 Conference of the North {A}merican
  Chapter of the Association for Computational Linguistics: Human Language
  Technologies, Volume 1 (Long and Short Papers)}, pages 4129--4138,
  Minneapolis, Minnesota. Association for Computational Linguistics.

\bibitem[{Huybregts(1984)}]{huybregts1984weak}
Riny Huybregts. 1984.
\newblock The weak inadequacy of context-free phrase structure grammars.
\newblock \emph{Van periferie naar kern}, pages 81--99.

\bibitem[{Joshi(1985)}]{joshi1985tree}
Aravind~Krishna Joshi. 1985.
\newblock \href
  {https://www.cs.sfu.ca/~anoop/courses/ReadingGroup-Summer-2006/joshi85.pdf}
  {Tree adjoining grammars: How much context-sensitivity is required to provide
  reasonable structural descriptions?}

\bibitem[{Kallmeyer(2010)}]{kallmeyer2010parsing}
Laura Kallmeyer. 2010.
\newblock \emph{Parsing beyond context-free grammars}.
\newblock Springer Science \& Business Media.

\bibitem[{Klein and Manning(2001)}]{klein2001parsing}
Dan Klein and Christopher~D Manning. 2001.
\newblock \href {https://aclanthology.org/P01-1044.pdf} {Parsing with treebank
  grammars: Empirical bounds, theoretical models, and the structure of the
  {P}enn treebank}.
\newblock In \emph{Proceedings of the 39th Annual Meeting of the Association
  for Computational Linguistics}, pages 338--345.

\bibitem[{Koopman and Szabolcsi(2000)}]{koopman2000verbal}
Hilda~Judith Koopman and Anna Szabolcsi. 2000.
\newblock \emph{Verbal complexes}.
\newblock 34. MIT Press.

\bibitem[{Li et~al.(2015)Li, Tarlow, Brockschmidt, and Zemel}]{li2015gated}
Yujia Li, Daniel Tarlow, Marc Brockschmidt, and Richard Zemel. 2015.
\newblock \href {https://arxiv.org/pdf/1511.05493.pdf} {Gated graph sequence
  neural networks}.
\newblock \emph{arXiv preprint arXiv:1511.05493}.

\bibitem[{Lin et~al.(2019)Lin, Tan, and Frank}]{lin2019open}
Yongjie~Lin Lin, Yi~Chern Tan, and Robert Frank. 2019.
\newblock \href {https://aclanthology.org/W19-4825.pdf} {Open sesame: Getting
  inside {BERT}’s linguistic knowledge}.
\newblock In \emph{Proceedings of the Second BlackboxNLP Workshop on Analyzing
  and Interpreting Neural Networks for NLP}.

\bibitem[{Liu et~al.(2019)Liu, Ott, Goyal, Du, Joshi, Chen, Levy, Lewis,
  Zettlemoyer, and Stoyanov}]{liu2019roberta}
Yinhan Liu, Myle Ott, Naman Goyal, Jingfei Du, Mandar Joshi, Danqi Chen, Omer
  Levy, Mike Lewis, Luke Zettlemoyer, and Veselin Stoyanov. 2019.
\newblock \href {https://arxiv.org/pdf/1907.11692.pdf} {Ro{BERT}a: A robustly
  optimized {BERT} pretraining approach}.
\newblock \emph{arXiv preprint arXiv:1907.11692}.

\bibitem[{Ljungl{\"o}f(2012)}]{ljunglof-2012-practical}
Peter Ljungl{\"o}f. 2012.
\newblock \href {https://aclanthology.org/W12-4617} {Practical parsing of
  parallel multiple context-free grammars}.
\newblock In \emph{Proceedings of the 11th International Workshop on Tree
  Adjoining Grammars and Related Formalisms ({TAG}+11)}, pages 144--152, Paris,
  France.

\bibitem[{Loshchilov and Hutter(2018)}]{loshchilov2018decoupled}
Ilya Loshchilov and Frank Hutter. 2018.
\newblock \href {https://openreview.net/pdf?id=Bkg6RiCqY7} {Decoupled weight
  decay regularization}.
\newblock In \emph{International Conference on Learning Representations}.

\bibitem[{Moortgat(1999)}]{moortgat1999meaningful}
Michael Moortgat. 1999.
\newblock \href
  {http://festschriften.illc.uva.nl/j50/contribs/moortgat/moortgat.pdf}
  {Meaningful patterns}.
\newblock \emph{Linguistics}, 210:4.

\bibitem[{Morrill et~al.(2007)Morrill, Valent{\'\i}n, and
  Fadda}]{morrill2007dutch}
Glyn Morrill, Oriol Valent{\'\i}n, and Mario Fadda. 2007.
\newblock \href
  {https://link.springer.com/chapter/10.1007/978-3-642-00665-4_22} {Dutch
  grammar and processing: A case study in {TLG}}.
\newblock In \emph{International Tbilisi Symposium on Logic, Language, and
  Computation}, pages 272--286. Springer.

\bibitem[{Muskens(2007)}]{muskens2007separating}
Reinhard Muskens. 2007.
\newblock \href
  {https://link.springer.com/content/pdf/10.1007/s11168-007-9035-1.pdf}
  {Separating syntax and combinatorics in categorial grammar}.
\newblock \emph{Research on language and computation}, 5(3):267--285.

\bibitem[{Pullum and Gazdar(1982)}]{pullum1982natural}
Geoffrey~K Pullum and Gerald Gazdar. 1982.
\newblock \href {https://link.springer.com/content/pdf/10.1007/BF00360802.pdf}
  {Natural languages and context-free languages}.
\newblock \emph{Linguistics and Philosophy}, 4(4):471--504.

\bibitem[{Rogers et~al.(2020)Rogers, Kovaleva, and
  Rumshisky}]{rogers-etal-2020-primer}
Anna Rogers, Olga Kovaleva, and Anna Rumshisky. 2020.
\newblock \href {https://doi.org/10.1162/tacl_a_00349} {A primer in
  {BERT}ology: What we know about how {BERT} works}.
\newblock \emph{Transactions of the Association for Computational Linguistics},
  8:842--866.

\bibitem[{Sang and De~Meulder(2003)}]{sang2003introduction}
Erik Tjong~Kim Sang and Fien De~Meulder. 2003.
\newblock Introduction to the conll-2003 shared task: Language-independent
  named entity recognition.
\newblock In \emph{Proceedings of the Seventh Conference on Natural Language
  Learning at HLT-NAACL 2003}, pages 142--147.

\bibitem[{Seki et~al.(1991)Seki, Matsumura, Fujii, and
  Kasami}]{seki1991multiple}
Hiroyuki Seki, Takashi Matsumura, Mamoru Fujii, and Tadao Kasami. 1991.
\newblock \href
  {https://www.sciencedirect.com/science/article/pii/030439759190374B/pdf?md5=a25be6b71683a69af87be6060f523d5d&pid=1-s2.0-030439759190374B-main.pdf}
  {On multiple context-free grammars}.
\newblock \emph{Theoretical Computer Science}, 88(2):191--229.

\bibitem[{Shieber(1985)}]{shieber1985evidence}
Stuart~M Shieber. 1985.
\newblock \href {https://apps.dtic.mil/sti/pdfs/ADA460998.pdf} {Evidence
  against the context-freeness of natural language}.
\newblock In \emph{Philosophy, language, and artificial intelligence}, pages
  79--89. Springer.

\bibitem[{Steedman(1985)}]{steedman1985dependency}
Mark Steedman. 1985.
\newblock \href {https://www.jstor.org/stable/pdf/414385.pdf} {Dependency and
  co{\"o}rdination in the grammar of dutch and english}.
\newblock \emph{Language}, pages 523--568.

\bibitem[{van~der Burgh and Verberne(2019)}]{vanderburgh2019merits}
Benjamin van~der Burgh and Suzan Verberne. 2019.
\newblock \href {http://arxiv.org/abs/1910.00896} {The merits of universal
  language model fine-tuning for small datasets -- a case with dutch book
  reviews}.

\bibitem[{Van~Noord et~al.(2013)Van~Noord, Bouma, Van~Eynde, De~Kok, Van~der
  Linde, Schuurman, Sang, and Vandeghinste}]{van2013large}
Gertjan Van~Noord, Gosse Bouma, Frank Van~Eynde, Daniel De~Kok, Jelmer Van~der
  Linde, Ineke Schuurman, Erik Tjong~Kim Sang, and Vincent Vandeghinste. 2013.
\newblock \href {https://doi.org/10.1007/978-3-642-30910-6_9} {Large scale
  syntactic annotation of written {Dutch}: Lassy}.
\newblock In \emph{Essential speech and language technology for {Dutch}}, pages
  147--164. Springer, Berlin, Heidelberg.

\bibitem[{Vilares et~al.(2020)Vilares, Strzyz, S{\o}gaard, and
  G{\'o}mez-Rodr{\'\i}guez}]{vilares2020parsing}
David Vilares, Michalina Strzyz, Anders S{\o}gaard, and Carlos
  G{\'o}mez-Rodr{\'\i}guez. 2020.
\newblock \href {https://ojs.aaai.org/index.php/AAAI/article/view/6446/6302}
  {Parsing as pretraining}.
\newblock In \emph{Proceedings of the AAAI Conference on Artificial
  Intelligence}, volume~34, pages 9114--9121.

\bibitem[{Wijnholds and Moortgat(2021)}]{wijnholds-moortgat-2021-sick}
Gijs Wijnholds and Michael Moortgat. 2021.
\newblock \href {https://doi.org/10.18653/v1/2021.eacl-main.126} {{SICK}-{NL}:
  A dataset for {D}utch natural language inference}.
\newblock In \emph{Proceedings of the 16th Conference of the European Chapter
  of the Association for Computational Linguistics: Main Volume}, pages
  1474--1479, Online. Association for Computational Linguistics.

\end{thebibliography}
\bibliographystyle{acl_natbib}

% \appendix

% \section{Example Appendix}
% \label{sec:appendix}

\end{document}